\documentclass[conference]{IEEEtran}
\IEEEoverridecommandlockouts

\usepackage{cite}
\usepackage{amsmath,amssymb,amsfonts}
\usepackage{graphicx}
\usepackage{textcomp}
\usepackage{enumitem}
\usepackage{xcolor}
\usepackage{tikz}
\usepackage{verbatim}
\usetikzlibrary{shapes.gates.logic.US,trees,positioning,arrows}

\usepackage{algorithmicx}
\usepackage[ruled]{algorithm}
\usepackage{algpseudocode}

\usepackage{multirow}

\usepackage{caption}
\usepackage{hyperref}
\usepackage{subcaption}
\usepackage{changepage}
\usepackage{lipsum}
\usepackage[left=1.8cm,right=1.8cm,top=2.4cm,bottom=1.8cm,bindingoffset=0cm]{geometry}

\def\BibTeX{{\rm B\kern-.05em{\sc i\kern-.025em b}\kern-.08em
    T\kern-.1667em\lower.7ex\hbox{E}\kern-.125emX}}
\begin{document}

\title{
\huge Node Templates to improve Reusability and Modularity of Behavior Trees 
\thanks{
$^1$%The author is with Italian Institute of Technology. 
Work done during an internship in Sberbank Robotics Laboratory. \newline
\texttt{evgenii.safronov@skoltech.ru}
}
}

\author{\IEEEauthorblockN{Evgenii Safronov$^1$}}

%
%\iffalse
%\and
%\IEEEauthorblockN{Michael Vilzmann}
%\and
%\IEEEauthorblockN{Konstantin Kondak}
%\and
%\IEEEauthorblockN{Dzmitry Tsetserukou}
%\fi

\newcommand{\mnt}{\emptyset}
\newcommand{\msR}{\mathbb{R}}
\newcommand{\msS}{\mathbb{S}}
\newcommand{\msF}{\mathbb{F}}

\newcommand{\nt}{$\emptyset$}
\newcommand{\sR}{$\mathbb{R}$}
\newcommand{\sS}{$\mathbb{S}$}
\newcommand{\sF}{$\mathbb{F}$}

\maketitle
\begin{abstract}
Behavior Trees (BTs) got the robotics society attention not least thanks to their modularity and reusability. The subtrees of BTs could be treated as separate behaviors and therefore reused. We address the following research question: do we exploit the full power of BT on these properties? We suggest to generalise the idea of subtree reuse to ``node templates'' concept, which allows to represent an arbitrary nodes collection. In addition, previously hardcoded behaviors such as Node* and many Decorator nodes could be implemented in a memory-based BT by node templates.
\end{abstract}

\iffalse
\fi
\section{Introduction}
Behaviors Tree (BT) is an architecture for task level control which came originally from computer games\cite{champandard2012behavior}. After adoption as a Non-Playable Character (NPC) control system, they acquired attention from the robotics community. They were suggested for different platforms e.g., unmanned aerial vehicles \cite{ogren2012increasing}\cite{my}, surgical robots \cite{surg}, robot assistants \cite{ass} and in various use cases e.g., manual tree designing \cite{SwedishBT} or learning control policy in a BT form \cite{Banerjee2018AutonomousAO}.  Modularity, scalability, and reusability are mentioned among the benefits of BTs \cite{book}. These properties influence the ability of end-user to implement complex mission scenarios in the behavior tree.  %Other well-known control architecture, Finite State Machine, is based on directed graph with unlimited finite possible number of incoming and outcoming edges. 
Modularity and reusability in programming languages are often accomplished by an incapsulation. For example, in object-oriented programming (OOP) objects are instances of classes, with possibly different arguments passed on the construction (instantiation). The closest known entity in a BT environment is a subtree, a collection of all children recursively starting from a certain node. Another popular approach is to prepare a set of decorators which incorporate desired behaviors. However, decorators in known to author works are hardcoded apriori and limited to one child. In this work, we extend the idea of subtree reuse to \emph{node templates} - an arbitrary nodes collection. Moreover, as number of nodes in each instantiated node template could vary according to defined rules, node templates reuse a certain ``\textit{behavior}'' rather than copy a collection of nodes. Here and after, by ``node template'' we mean the reused abstract definition similar to ``class'' concept in OOP, ``templated node'' refers to the instantiation of node template analogous to an ``object''. Finally, the predefined node classes are named ``primary''. For this work we use 6 primary node types: Sequence, Selector, Skipper, Parallel, Action and Condition, as defined in \cite{my}. 
\par 
In this work, the BT is defined over the memory layer\cite{my}. However, we follow the common top-down periodical tick execution, which is explained in an introduction to BTs\cite{book}. The execution involves the traversal of the tree structure.% where leaf nodes involves interaction with the environment (blackboard-like memory, in our case), while non-leaf nodes defines the logic of execution.
As a child node in a tree connected to a parent by the only one edge, subtrees in BT are naturally separated and could be treated as independent behaviors. Reuse of subtrees is widely supported by BT frameworks \cite{SwedishBT}. 
\section{Motivation}
Incorporating frequently used behaviors as subtrees, in addition, enhances scalability and readability properties. However, the reusability of subtrees is limited. We might need a slightly different behavior over a similar subtree structure. Behavior alternation could be done by blackboard variables assignment prior to entering the subtree execution. A. Shoulson \cite{parbt} introduced the parametrization of subtrees to avoid possible mistakes with the blackboard data flow. Still, subtrees could not be reused intensively with different leaf nodes or between different projects and platforms. Another noticeable feature of other BT definitions is the existence of extra control nodes such as decorators or ``Node*'' (or ``Control node with memory''). They were often excluded from theoretical analysis and from works with tree generation but were stated as useful in practice. All mentioned above motivated us to extend the idea of subtree reuse. The idea is to reuse not a subtree but a collection of nodes with possibly vacant children places as arguments. Arguments are not limited to node names e.g., they could be parameters for actions (explained in Section III). To avoid name conflicts within internal variables used in a node template each node is forced to have a unique name. In addition, it improves general readability of a tree, as the name of a node hints its purpose.
\begin{figure*}[ht]
    \centering
\begin{subfigure}{0.25\textwidth}
\includegraphics[width=0.99\textwidth]{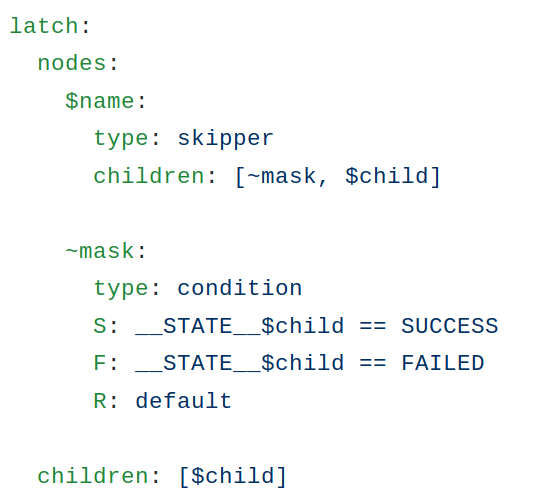}
\caption{\footnotesize  The Latch node template definition. ``\$'' operator substitutes the following argument. ``\$name'' or ``\texttildelow'' substitutes a name of the templated node.}
\label{fig:latch-def}
\end{subfigure}
~
\begin{subfigure}{0.21\textwidth}
\includegraphics[width=0.99\textwidth]{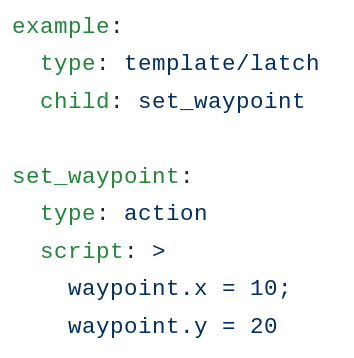}
\caption{\footnotesize The templated node description for the tree construction.}
\label{fig:latch-ins}
\end{subfigure}
~
\begin{subfigure}{0.15\textwidth}
\includegraphics[width=0.99\textwidth]{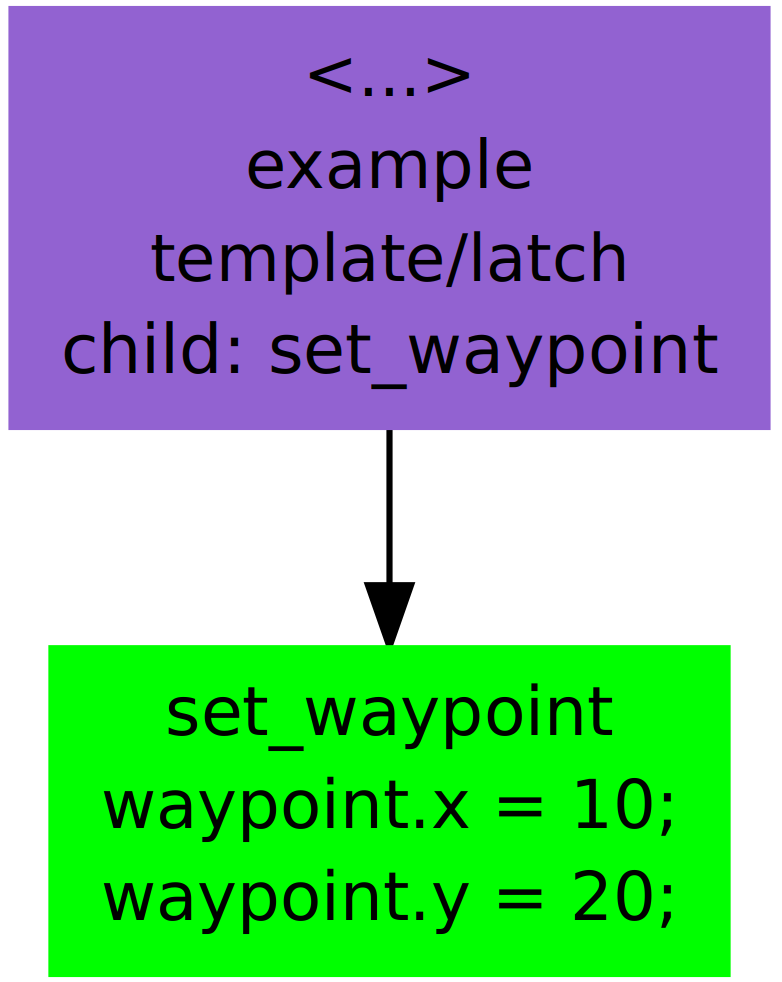}
\caption{\footnotesize User view on the tree in case the templated node is seen as a control node of a new type.}
\label{fig:latch-hidden}
\end{subfigure}
~
\begin{subfigure}{0.31\textwidth}
\includegraphics[width=0.99\textwidth]{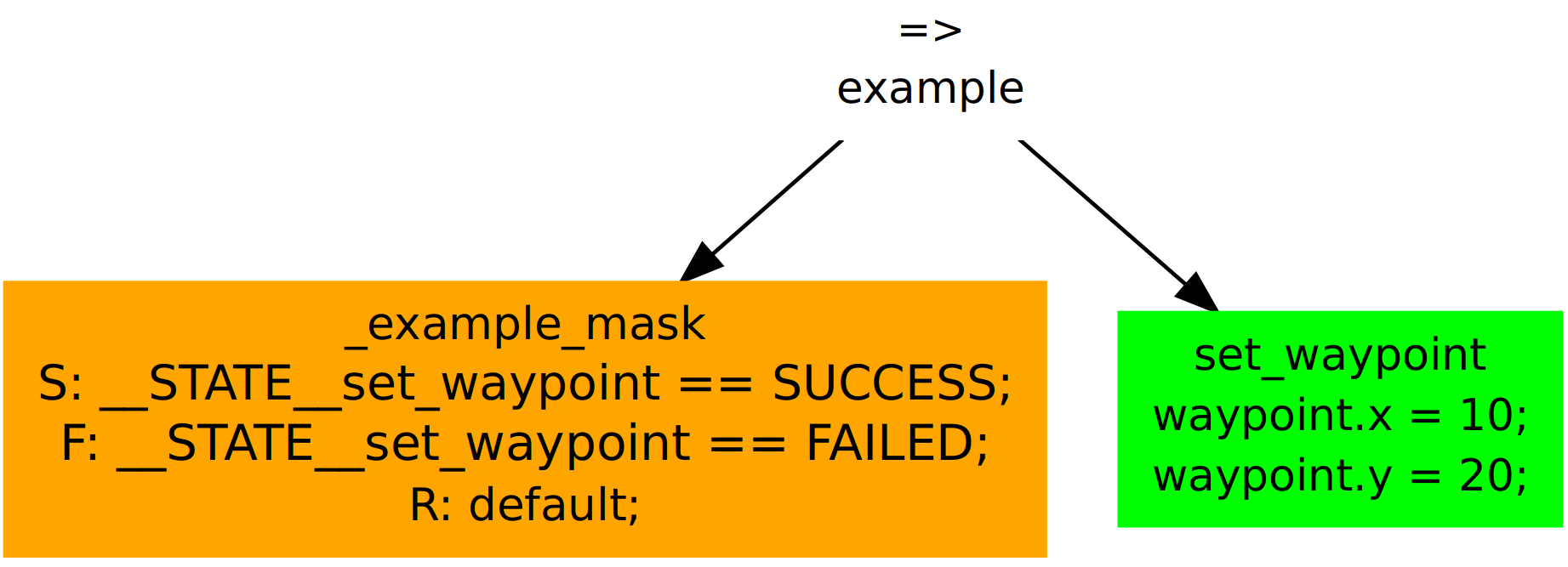}
\caption{\footnotesize An expanded tree that reveals substituted nodes and the child. ``\_\_STATE\_\_'' prefix stands for the variable which remembers last tick returns.}
\label{fig:latch-exp}
\end{subfigure}
\caption{\footnotesize A simpler version of ``Latch'' templated node explained. YAML-based tree description file contains the definition of the template ``Latch'' (\ref{fig:latch-def}) and nodes in the tree (\ref{fig:latch-ins}). After building, the tree consists of templated node ``example'' of type ``latch'' (\ref{fig:latch-hidden}) while in fact, it is a collection of nodes (\ref{fig:latch-exp}). A symbol $=>$  denotes a Skipper node\cite{my}, action and condition nodes are highlighted with green and orange colors respectively.} 
\label{fig:latch_example}
\end{figure*}

\section{Definition} To define ``node template'' we have to discuss the following topics:
\begin{itemize}
	\item the view of the templated node from the user side
	\item the algorithm of building tree nodes from the node template instantiation
	\item the rules of a node template definition
\end{itemize}
\paragraph{User view} 
From the user side the templated node looks like a node in a tree. If the node template accepts other node names as arguments, then the user could treat it is a control node. Otherwise, the templated node would act as a leaf node. Notice that in this case node template is an analog of subtree with parameters. Therefore, \textit{node templates generalize subtrees with parameters}.
\paragraph{Defining a node template} 
Without loss of generality, we assume the tree could be constructed completely from a text description. Indeed, control nodes could be described by their name, children list, and type, while leaf nodes definition in a worst case could be associated with their code in some programming language. Node template definition contains the algorithm which replaces the templated node with a collection of nodes with substituted arguments. In general, the node template's arguments substitution and therefore definition could be done in many ways e.g., involving as a separate script. In this paper, we assume a simpler scenario where we only have certain rules for the arguments substitution in a text template. The arguments could be either scalar or list of scalars.

\paragraph{Building a templated node}
To build a templated node, we would substitute arguments in a node template text definition and then treat this text as a definition of a collection of nodes. If node template definition contains another templated node, it would be recursively expanded according to its definition and forwarded arguments. If one of the arguments is a list, we can define a nodes collection which would be reproduced with iteration over the list. Refer to the Examples section for a better understanding of this part.
\section{Examples}

\paragraph{Latch} Let's take a look at an example of Latch decorator (Fig. \ref{fig:latch_example}). It was suggested as a standalone hardcoded node to solve the waypoint sequence problem\cite{KlocknerStateful}. If we allow conditions to look at last evaluated states of each node as variables, then the implementation of Latch is extremely simple (Fig. \ref{fig:latch-def}, \ref{fig:latch-ins}). Otherwise, it requires a few more nodes\cite{my}. Nevertheless, all these nodes are going to be hidden under a template mask (Fig. \ref{fig:latch-hidden}), which accepts one argument, name of the child node. Reset node could be implemented in the same manner.
\paragraph{Node*}
\begin{figure}[h]
\includegraphics[width=0.45\textwidth]{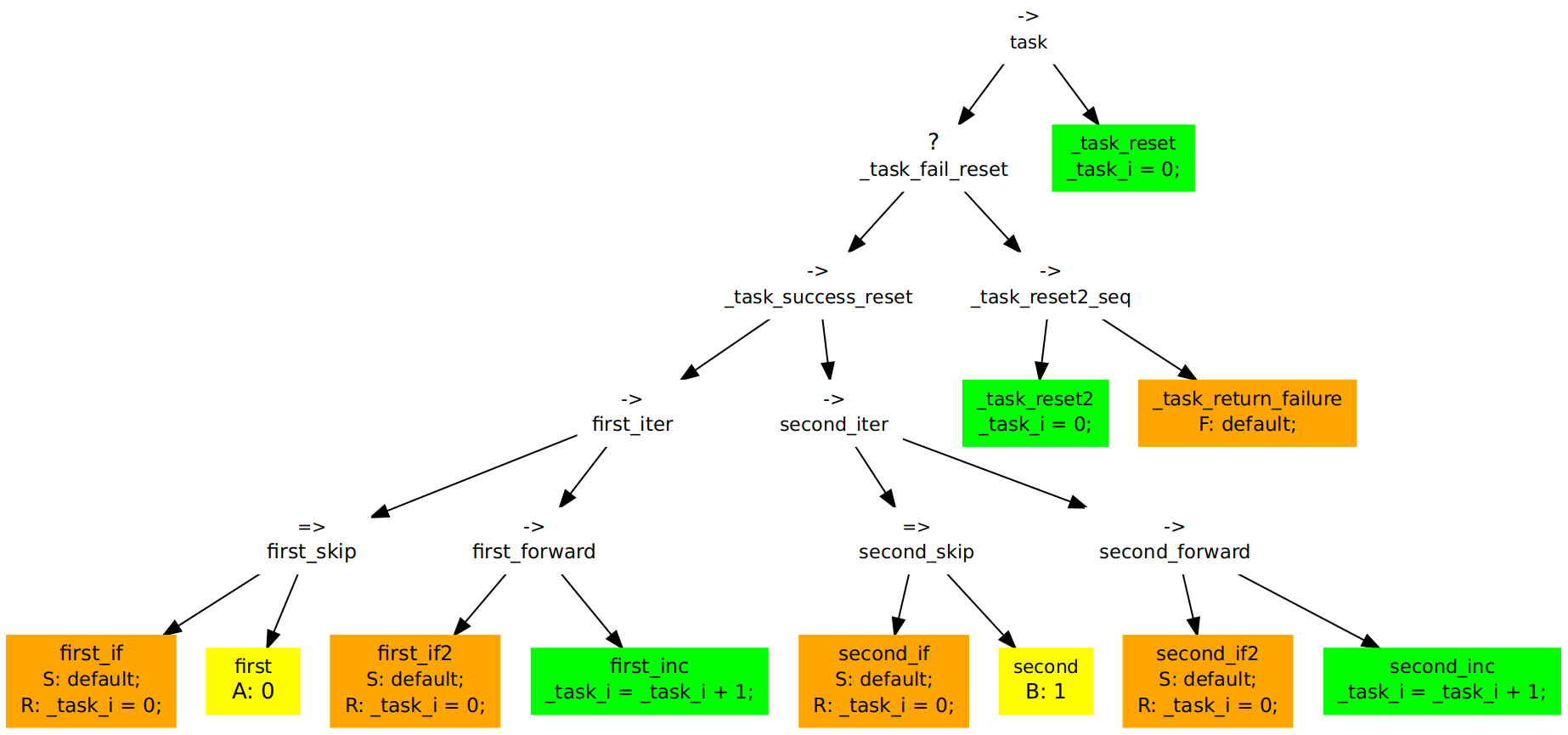}
\caption{\footnotesize A expanded \texttt{task} ``Sequence*'' templated node with two children. Templated node's children are highlighted in yellow to distinguish from other nodes.}
\label{fig:node_star}
\end{figure}
Another approach to implement ``waypoint mission'' behavior is a ``Control node with memory'' or so-called ``Node*'' family that was introduced in \cite{SwedishBT}. If a child returns \sS~state to the parent Sequence node (or more generally a ``return state'' category \cite{my}), then the parent remembers this result and do not reevaluate the child in later tick execution. Instead of populating of primary node types, we can create a node template which  accepts a variable number of children. 
\par These two and the other examples are available online as a part of ABTM library developed by the author. \footnote{\href{https://github.com/safoex/abtm}{https://github.com/safoex/abtm}}

\section{Conclusion}
Subtree reuse and decorator nodes were generalized with a novel node templates notion. The description of the templated definition and instantiation was provided in the paper. Two examples of common Latch decorator and Node* concepts are implemented to show the expressive power of node templates. Node templates improve readability hence fewer mistakes are made during BT design.
\par For further research we expect an application of node templates for BT generation.

\section{Acknowledgements}
We thank Michele Colledanchise for his remarks on this paper and all colleagues in Sberbank Robotics Laboratory for their hospitality during the internship.

\bibliography{iros2019}
\bibliographystyle{ieeetr}
\end{document}